%% file: root.tex
\documentclass[letterpaper, 10 pt, journal, twoside]{ieeetran}
\IEEEoverridecommandlockouts                              

\PassOptionsToPackage{dvipsnames}{xcolor}
\PassOptionsToPackage{table}{xcolor}

\input{macros/macros.tex}

\usepackage{amsmath}
\usepackage{amssymb}
\usepackage{booktabs}
\usepackage{makecell}
\usepackage[ruled,noend,linesnumbered]{algorithm2e}
\usepackage{titlesec}
\usepackage{wrapfig}
\usepackage{array}
\usepackage{multirow}
\usepackage{colortbl}
\usepackage{rotating}
\usepackage{makecell}
\usepackage{tabularx}
\usepackage{soul}

\usepackage{grffile}
\usepackage{xspace}
\usepackage{macros/mysymbols}

\usepackage{comment} 
\usepackage{url}
\usepackage{paralist}

\usepackage[belowskip=0pt,aboveskip=5pt,font=small]{caption}
\usepackage[belowskip=0pt,aboveskip=0pt,font=small]{subcaption}

\title{Bi-directional Domain Adaptation for \\ Sim2Real Transfer of Embodied Navigation Agents}

\author{Joanne Truong$^{1}$ and Sonia Chernova$^{1,2}$ and Dhruv Batra$^{1,2}$
\thanks{Manuscript received: October, 16, 2020; Revised January, 15, 2021; Accepted February, 16, 2021.}
\thanks{This paper was recommended for publication by Editor Eric Marchand upon evaluation of the Associate Editor and Reviewers' comments.

The Georgia Tech effort was supported in part by NSF, AFRL, DARPA, ONR YIPs, ARO PECASE. JT was supported by an NSF GRFP and a Google Women Techmaker's Fellowship.} 
\thanks{$^{1}$JT, SC, and DB are with Georgia Institute of Technology
        {\tt\footnotesize \{truong.j, chernova, dbatra\}@gatech.edu}}%
\thanks{$^{2} $SC and DB are with Facebook AI Research
        {\tt\footnotesize \{schernova, dbatra\}@fb.com}}%
\thanks{Digital Object Identifier (DOI): see top of this page.}
}

\begin{document}
\bstctlcite{IEEEexample:BSTcontrol}

\maketitle

\input{sections/0_abstract}

\input{sections/1_introduction}

\input{sections/2_BDA}

\input{sections/3_experimental_setup}

\input{sections/5_experiments}

\input{sections/6_related_work}

\input{sections/7_conclusion}

\input{sections/8_acknowledgements}


{
\bibliographystyle{style/IEEEtran}
\bibliography{bib/strings,bib/main}
}


\end{document}

%% file: macros/macros.tex
\newcommand{\xhdr}[1]{\vspace{2pt}\noindent\textbf{#1}}

\newcommand{\todo}[1]{\textcolor{red}{TODO: #1}}%
\newcommand{\reffig}[1]{Fig.~\ref{#1}}

\newcommand{\reftab}[1]{Table ~\ref{#1}}

\newcommand{\oa}{$\mathcal{OA}$\xspace}%
\newcommand{\da}{$\mathcal{DA}$\xspace}%

\newcommand{\forwardcyc}{$\calI^{\text{real}}\to \mathcal{G}_{sim}(\calI^{\text{real}}) \to \mathcal{G}_{\text{real}}(\mathcal{G}_{\text{sim}}(\calI^{\text{real}})) \approx \calI^{\text{real}}$ \xspace}%
\newcommand{\backwardcyc}{$\calI^{\text{sim}}\to \mathcal{G}_{\text{real}}(\calI^{\text{sim}})\to \mathcal{G}_{\text{sim}}(\mathcal{G}_{\text{real}}(\calI^{\text{sim}}))\approx \calI^{\text{sim}}$}%

\newcommand{\pis}{$\pi^{100M}_{\text{Gibson}^{\text{no\_noise}}}$\xspace}%
\newcommand{\pit}{$\pi^{100M}_{\text{Gibson}^{O+D}}$\xspace}%
\newcommand{\pigt}{$\pi^{{1M}}_{\text{Gibson}^{O+D}}$\xspace}%
\newcommand{\pir}{$\pi^{{\text{BDA}-5k}}_{\text{Gibson}^{OA+DA}}$\xspace}%
\newcommand{\piroa}{$\pi^{{\text{BDA}-5k}}_{\text{Gibson}^{OA}}$\xspace}%
\newcommand{\pirda}{$\pi^{{\text{BDA}-5k}}_{\text{Gibson}^{DA}}$\xspace}%

\newcommand{\pisim}{$\pi_{\text{sim}}$\xspace}%

\newcommand{\labnn}{$\text{LAB}^{\text{no\_noise}}$\xspace}%
\newcommand{\labsens}{$\text{LAB}^{O}$\xspace}%
\newcommand{\labact}{$\text{LAB}^{D}$\xspace}%
\newcommand{\labn}{$\text{LAB}^{O+D}$\xspace}%

\newcommand{\gibnn}{$\text{Gibson}^{\text{no\_noise}}$\xspace}%
\newcommand{\gibn}{$\text{Gibson}^{O+D}$\xspace}%

\newcommand{\isim}{$I^{\text{sim}}$\xspace}%
\newcommand{\ireal}{$I^{\text{real}}$\xspace}%

\newcommand{\isimd}{$\{{I^{\text{sim}}}^N_{i=1}\}$\xspace}%
\newcommand{\ireald}{$\{{I^{\text{real}}}^N_{i=1}\}$\xspace}%
\newcommand{\imgon}{$\mathcal{I}^{O}$\xspace}%
\newcommand{\imgnn}{$\mathcal{I}^{\text{no\_noise}}$\xspace}%
\newcommand{\imggaus}{$\mathcal{I}^{\text{Gaussian}}$\xspace}%

\newcommand{\imgtsim}{$\mathcal{I}^{\text{sim}}_{t}$\xspace}%
\newcommand{\imgtreal}{$\mathcal{I}^{\text{real}}_{t}$\xspace}%
\newcommand{\streal}{$s^{\text{real}}_{t}$\xspace}%
\newcommand{\atreal}{$a^{\text{real}}_{t}$\xspace}%
\newcommand{\sreald}{$\{{s^{\text{real}}}^N_{i=1}\}$\xspace}%
\newcommand{\areald}{$\{{a^{\text{real}}}^N_{i=1}\}$\xspace}%
\newcommand{\simDA}{$\text{Sim}^{\mathcal{DA}}$\xspace}%
\newcommand{\pisimDA}{$\pi_{\text{Sim}^\mathcal{{DA}}}$\xspace}%
\newcommand{\pisimODA}{$\pi_{\text{Sim}^{\mathcal{OA} + \mathcal{DA}}}$\xspace}%

\newcommand{\ouralgfull}{Bi-directional Domain Adaptation\xspace}%
\newcommand{\ouralg}{BDA\xspace}%

\newcommand{\tableTitle}[1]{\normalsize{#1}}%
\newlength{\figwidth}%



%% file: sections/0_abstract.tex

\begin{abstract} Deep reinforcement learning models are notoriously data hungry, yet real-world data is expensive and time consuming to obtain. The solution that many have turned to is to use simulation for training before deploying the robot in a real environment. Simulation offers the ability to train large numbers of robots in parallel, and offers an abundance of data. However, no simulation is perfect, and robots trained solely in simulation fail to generalize to the real-world, resulting in a ``sim-vs-real gap". How can we overcome the trade-off between the abundance of less accurate, artificial data from simulators and the scarcity of reliable, real-world data? In this paper, we propose \ouralgfull (\ouralg), a novel approach to bridge the sim-vs-real gap in both directions-- \emph{real2sim} to bridge the visual domain gap, and \emph{sim2real} to bridge the dynamics domain gap. We demonstrate the benefits of BDA on the task of PointGoal Navigation. \ouralg with only 5k real-world (state, action, next-state) samples matches the performance of a policy fine-tuned with $\sim$600k samples, resulting in a speed-up of $\sim$120$\times$.  

\end{abstract}




%% file: sections/1_introduction.tex
\section{Introduction}



\IEEEPARstart{D}{eep} reinforcement learning (RL) methods have made tremendous progress in many high-dimensional tasks, such as navigation~\cite{ddppo}, manipulation~\cite{andrychowicz2020learning}, and locomotion~\cite{haarnoja2018learning}. 
Since RL algorithms are data hungry, and training robots in the real-world is slow, expensive, and difficult to reproduce, 
these methods are typically trained in simulation (where gathering experience is scalable, safe, cheap, and reproducible) before being deployed in the real-world. 

However, no simulator perfectly replicates reality. Simulators fail to model many aspects of the robot and the environment 
(noisy dynamics, sensor noise, wear and-tear, battery drainage, \etc). 
In addition, RL algorithms are prone to overfitting -- \ie, they learn to achieve strong performance in the environments they were trained in, 
but fail to generalize to novel environments.  
On the other hand, humans are able to quickly adapt to small changes in their environment.
The ability to quickly adapt and transfer skills is a key aspect of intelligence that we hope to reproduce in artificial agents. 

This raises a fundamental question -- How can we leverage imperfect but useful simulators to train robots while ensuring that the learned skills generalize to reality? 
This question is studied under the umbrella of `sim2real transfer' and has been a topic of much interest in the community 
\cite{graspgan, golemo2018sim, rcann, peng2018sim, tobin2017domain, yu2020learning, vrgoogles}. 

In this work, we first reframe the sim2real transfer problem into the following question -- 
\emph{given a cheap abundant low-fidelity data generator (a simulator) and an expensive scarce high-fidelity data source (reality), 
how should we best leverage the two to maximize performance of an agent in the expensive domain (reality)?} 
The status quo in machine learning is to pre-train a policy in simulation using large amounts of simulation data (potentially with domain 
randomization~\cite{tobin2017domain}) and then fine-tune this policy on the robot using the small amount of real data. Can we do better? 

\begin{figure*}
  \vspace{7 pt}
  \centering%
  \resizebox{\textwidth}{!}{
  \renewcommand{\tableTitle}[1]{\large{#1}}%
  \setlength{\tabcolsep}{1.5pt}%
  \renewcommand{\arraystretch}{0.8}%
  \renewcommand\cellset{\renewcommand\arraystretch{0.8}%
  \setlength\extrarowheight{0pt}}%
  \hspace{-0.25cm}\begin{tabular}{c c}
   \makecell{\includegraphics[width=0.7\textwidth]{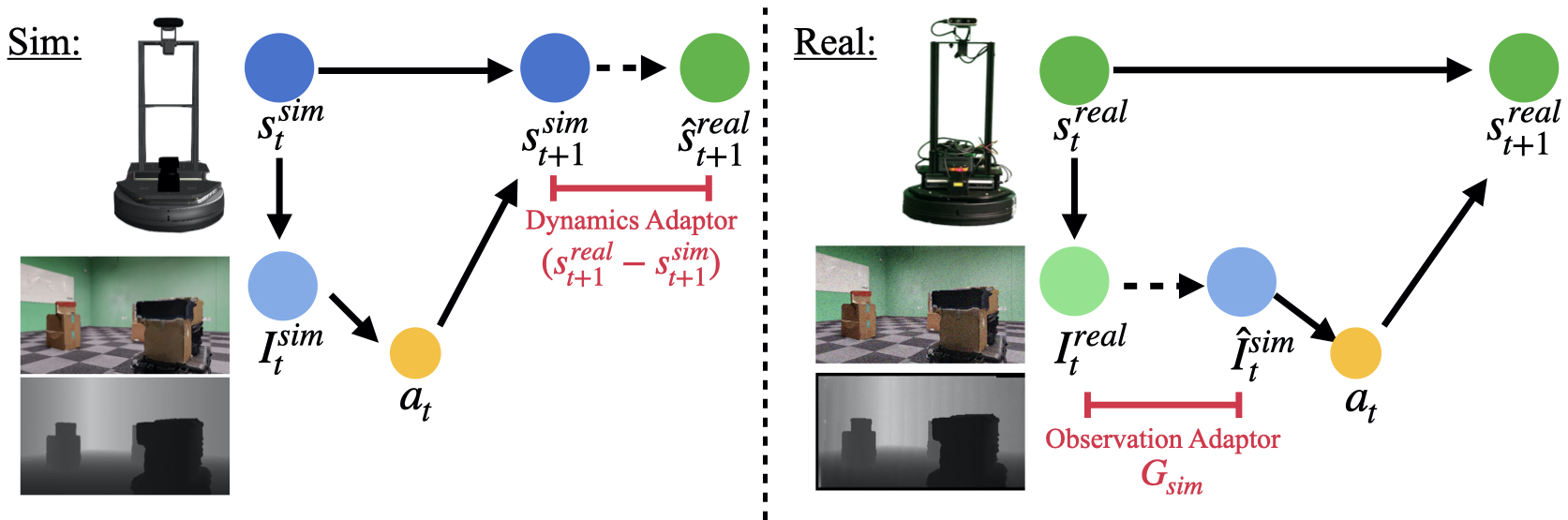} }&%
   \makecell{\includegraphics[width=0.3\textwidth]{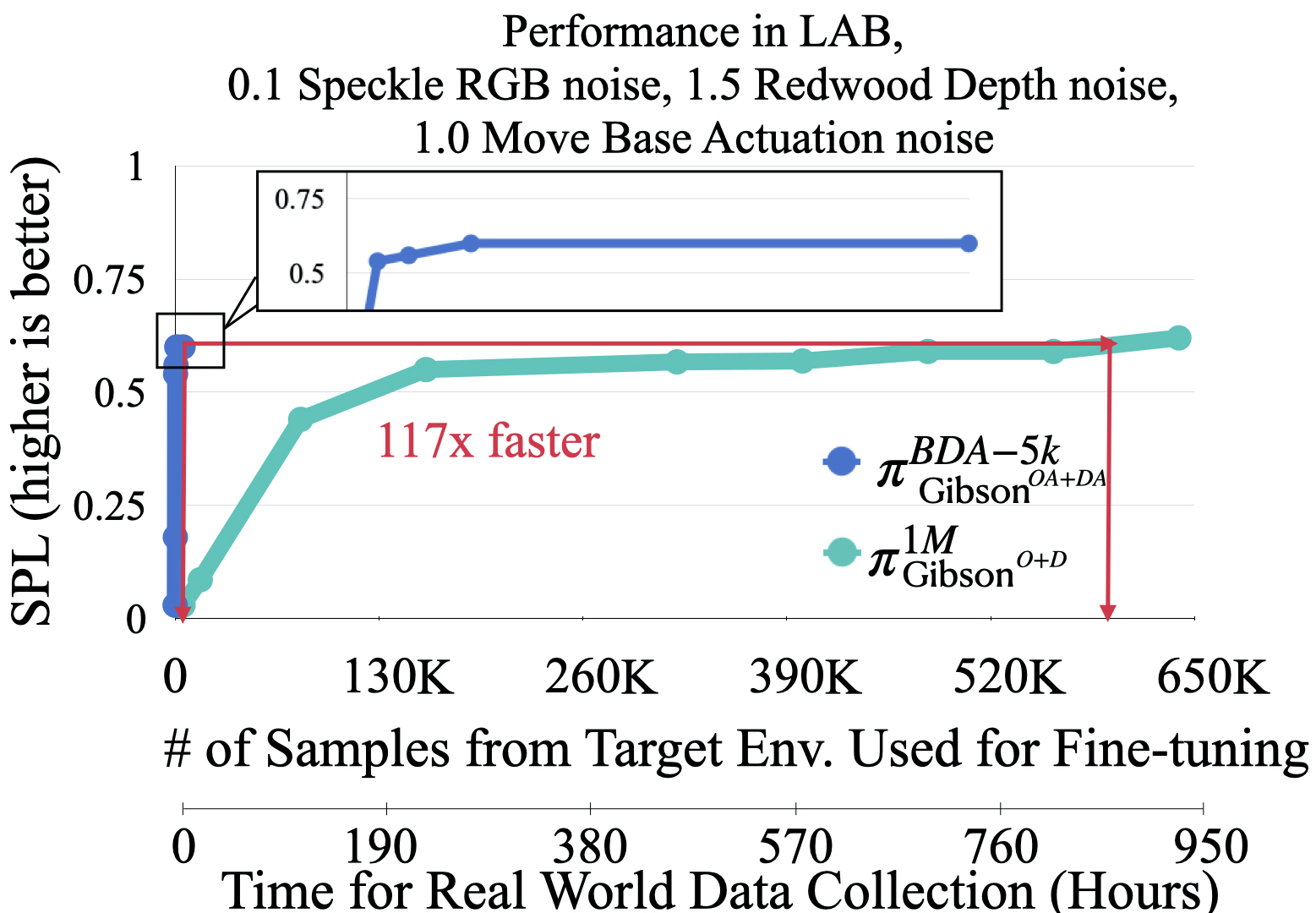}}\\
   (a) & (b)
  \end{tabular}}
  \caption{(a) Left: We learn a \emph{sim2real} dynamics adaptation module to predict residual errors between state transitions in simulation and reality. Right: We learn a  \emph{real2sim} observation adaptation module to translate images the robot sees in the real-world at test time to images that more closely align with what the robot has seen in simulation during training. (b) Using BDA, we achieve the same SPL as a policy finetuned directly in reality while using 117$\times$ less real-world data.}
  \label{fig:teaser}
  \vspace{-15pt}
  \end{figure*}

We contend that the small amount of expensive, high-fidelity data from reality is better utilized to adapt the simulator (and reduce the sim-vs-real gap) than to directly adapt the policy. Concretely, we propose \ouralgfull (\ouralg) between simulation and reality to answer this question. 
\ouralg reduces the sim-vs-real gap in two different directions (shown in \figref{fig:teaser}).

First, for sensory observations (\eg an RGB-D camera image $I$) we train a \emph{real2sim} observation adaptation module 
$\calO\calA: \calI^{\text{real}} \mapsto \calI^{\text{sim}}$.   
This can be thought of as `goggles' ~\cite{vrgoogles, xia2018gibson} that the agent 
puts on at deployment time to make real observations `look' like the ones seen during training in simulation. 
At first glance, this choice may appear counter-intuitive (or the `wrong' direction). 
We choose real2sim observation adaption instead of sim2real because this decouples sensing and acting. 
If the sensor characteristics in reality change but the dynamics remain the same (\eg same robot different camera), 
the policy does not need to be retrained, but only equipped with a re-trained observation adaptor. 
In contrast, changing a sim2real observation adaptor results in the generated observations being out of distribution for the policy, 
requiring expensive re-training of the policy. Our real2sim observation adaptor is based on CycleGAN~\cite{CycleGAN2017}, and thus 
does \emph{not} require any sort of alignment or pairing between sim and real observations, which can be prohibitive. 

Second, for transition dynamics $\calT: Pr(s_{t+1} \mid s_t, a_t)$ (the probably of transitioning from state $s_t$ to $s_{t+1}$ upon taking action $a_t$), 
we train a \emph{sim2real} dynamics adaptation module 
$\calD\calA: \calT^{\text{sim}} \mapsto \calT^{\text{real}}$. This can be thought of as a neural-augmented simulator \cite{golemo2018sim} 
or a specific kind of boosted ensembling method~\cite{schapire_ijcai99} -- where a simulator first makes predictions about state transitions 
and then a learned neural network predicts the residual between the simulator predictions and the state transitions observed in reality. 
At each time $t$ during training in simulation, $\calD\calA$ resets the simulator state from 
$s^{\text{sim}}_{t+1}$ (where the simulator believes the agent should reach at time $t+1$) to $\hat{s}^{\text{real}}_{t+1}$ (where $\calD\calA$ predicts the agent will reach in reality), thus exposing the policy to trajectories expected in reality. 
We choose sim2real dynamics adaptation instead of real2sim because this nicely exploits the fundamental asymmetry between the two domains -- 
simulators can (typically) be reset to arbitrary states, reality (typically) cannot. Once an agent acts in the real-world, it doesn't matter 
what corresponding state it would have reached in simulator, reality cannot be reset to it. 

Once the two modules are trained, \ouralg trains a policy in a simulator augmented with the dynamics adaptor ($\calD\calA$) 
and deploys the policy augmented with the observation adaptor ($\calO\calA$) to reality. 
This process is illustrated in \reffig{fig:teaser}a, left showing policy training in simulation and right showing its deployment in reality. 


We instantiate and demonstrate the benefits of \ouralg on the task of PointGoal Navigation (PointNav) \cite{anderson2018evaluation}, which involves 
an agent navigating in a previously unseen environment from a randomized initial starting location to a goal location specified in relative coordinates. 
For controlled experimentation, and due to COVID-19 restrictions, we use Sim2Sim transfer of PointNav policies 
as a stand-in for Sim2Real transfer. We conduct experiments in photo-realistic 3D simulation environments using 
Habitat-Sim \cite{habitat19iccv}, which prior work \cite{srcc} has found to have high \emph{sim2real predictivity}, 
meaning that inferences drawn in simulation experiments have a high likelihood of holding in reality on Locobot mobile robot \cite{locobot}.

In our experiments, we find that \ouralg is \emph{significantly} more sample-efficient than the baseline of fine-tuning a policy. 
Specifically, \ouralg trained on as few as 5,000 samples (state, action, next-state) from reality (equivalent of 7 hours to collect data in reality) is able to match the performance 
of baseline trained on \emph{585,000} samples from reality (equivalent of 836 hours to collect data in reality, or 3.5 months at 8 working hours per day), a speed-up of 117$\times$ (\reffig{fig:teaser}b).

While our experiments are conducted on the PointNav task, we believe our findings, and the core idea of \ouralgfull, is 
broadly applicable to a number of problems in robotics and reinforcement learning.

%% file: sections/2_BDA.tex
\section{\ouralgfull(\ouralg)}
\label{sec:bda}

We now describe the two key components of \ouralgfull(\ouralg) in detail -- (1) \emph{real2sim} observation adaptation module \oa to close the visual domain gap, and (2) \emph{sim2real} dynamics adaptation module \da to close the dynamics domain gap. 

\xhdr{Preliminaries and Notation.} We formulate our problem by representing both the source and target domain as a Markov Decision Process (MDP). A MDP is defined by the tuple ($\mathcal{S}, \mathcal{A}, \mathcal{T}, \mathcal{R}, \gamma$), where $s \in \mathcal{S}$ denotes states, $a \in \mathcal{A}$ denotes actions, $\mathcal{T}(s,a,s') = Pr(s' \mid s,a)$ is the transition probability, $\mathcal{R}: \mathcal{S} \times \mathcal{A} \rightarrow \mathbb{R}$ is the reward function, and $\gamma$ is a discount factor. In RL, the goal is to learn a policy $\pi: \mathcal{S} \rightarrow \mathcal{A}$ to maximize expected reward.

\subsection{System Architecture}
\label{sec:domain_adapt}
\vspace{-10pt}
\IncMargin{1em}
\begin{algorithm}
\SetAlgoNoLine%
Train behavior policy \pisim in Sim\\
\For{$t$ = 0, ..., N \text{steps}}{
 Collect \imgtsim $\sim$  Sim rollout (\pisim)\\
 Collect \imgtreal, \streal, \atreal $\sim$  Real rollout (\pisim) \\
}
Train \oa(\isimd, \ireald)\\
Train \da(\sreald, \areald)\\
\simDA $\leftarrow$ Augment Source with \da \\
\For{$j$ = 0, ..., K \text{steps}}{
 \pisimDA $\leftarrow$ Finetune \pisim in \simDA
}
\pisimODA $\leftarrow$ Apply \oa at test-time\\
Test \pisimODA in Real 
 \caption{Bi-directional Domain Adaptation}
 \label{alg:architecture}
\end{algorithm}
\vspace{-10pt}
\DecMargin{1em}


%

\xhdr{Observation Adaptation.}
We consider a \emph{real2sim} domain adaptation approach to deal with the visual domain gap. 

We leverage CycleGAN \cite{CycleGAN2017}, a pixel-level image-to-image translation technique that uses a cycle-consistency loss function with unpaired images. We start by using a behavior policy \pisim trained in simulation to sample rollouts in simulation and reality to collect RGB-D images \imgtsim and \imgtreal at time $t$ (line 3). The dataset of N unpaired images \isimd and \ireald is used to train \oa, to translate \isimd $\mapsto$ \ireald (line 5). 
\oa learns a mapping $\mathcal{G}_{\text{sim}}: \calI^{\text{real}} \mapsto \calI^{\text{sim}}$, an inverse mapping $\mathcal{G}_{\text{real}}: \calI^{\text{sim}} \mapsto \calI^{\text{real}}$, and adversarial discriminators $\mathcal{D}_{\text{real}}$, $\mathcal{D}_{\text{sim}}$. Although our method focuses on adaptation from \emph{real2sim}, learning both mappings encourages the generative models to remain cycle-consistent, i.e., forward cycle: \forwardcyc and backwards cycle: \backwardcyc. The ability to learn mappings from unpaired images from both domains is important because it is difficult to accurately collect paired images between simulation and reality.

A \emph{real2sim} approach for adapting the visual domain offers many advantages over a \emph{sim2real} approach because it disentangles the sensor adaptation module from our policy training. This enables us to remove an additional bottleneck during the RL policy training process; we can train \oa in parallel with the RL policy, thus reducing the overall training time needed. In addition, if the sensor observation noise in the environment changes, the base policy can be kept frozen, and only \oa will have to be retrained.  

\xhdr{Dynamics Adaptation.}
To close the dynamics domain gap, we follow a \emph{sim2real} approach. 
Starting with the behavior policy \pisim, we collect state-action pairs (\streal, \atreal) in the real-world (line 4). The state-action pairs are used to train \da, a 3 layer multilayer perceptron regression network, that learns the residual error between the state transitions in simulation and reality $\calT^{\text{sim}} \mapsto \calT^{\text{real}}$ (line 7). Specifically, \da learns to estimate the change in position and orientation $\Delta s^{\text{real}}$ : ($s^{\text{real}}_{t+1} - s^{\text{real}}_t)$. We use a weighted MSE loss function,  $\frac{1}{n}\sum_{n=1}^{N}\mathbf{w}^\top(\Delta s^{\text{real}}_{n}- \Delta \hat{s}^{\text{real}}_{n})^2$. 
For our experiments, the state \streal = ($x^{\text{real}}_t$, $y^{\text{real}}_t$, $\theta^{\text{real}}_t$), is represented by the position and orientation of the robot at timestep $t$. We placed twice as much weight on the prediction terms for the robot's position than for its orientation because getting the position correct is more important for our performance metric. 
Once trained, \da is used to augment the source environment (line 7). We finetune \pisim in the augmented simulator, \simDA (lines 8-9). Our hypothesis (which we validate in our experiments) is using real-world data to adapt the simulator via our \da model pays off because we can then train RL policies in this \da-augmented simulator for large amounts of experience cheaply.  We use \oa at test time (line 10). Finally, we test our policy trained with BDA in the real-world (line 11). 
To recap, \ouralg has a number of advantages over the status quo (of directly using real data to fine-tune a simulation
trained policy) that we demonstrate in our experiments: 
\begin{inparaenum}[(1)]
\item Decouples sensing and acting, 
\item Does not require paired training data, 
\item The data to train both modules can be collected jointly (by gathering experience from a behavior policy in reality) 
but the two 
can be trained in parallel independently of each other,  
\item Similar to model-based RL~\cite{sutton2018reinforcement}, reducing the sim-vs-real gap is made significantly more sample-efficient than directly fine-tuning the policy. 
\end{inparaenum}

%% file: sections/3_experimental_setup.tex
\section{Experimental Setup: Sim2Sim Transfer for Point-Goal Navigation}
\label{sec:setup}
Our goal in this work is to enable sample efficient Sim2Real transfer for the task of PointGoal Navigation (PointNav) ~\cite{anderson2018evaluation}. However, for controlled experiments and due to COVID-19 restrictions, we study Sim2Sim transfer as a stand-in for Sim2Real. Specifically, we train policies in a ``source" simulator (which serves as `Sim' in `Sim2Real') and transfer it to a ``target" simulator (which serves as `Real' in `Sim2Real'). We add observation and dynamics noise to the target simulator to mimic the noise observed in reality. Notice that these noise models are purely for the purpose of conducting controlled experiments and are not available to the agent (which must adapt and learn from samples of state and observations). Since no noise model is perfect (just like no simulator is perfect), we experiment with a range of noise models and report results with multiple target simulators. Our results show consistent improvements regardless of the noise model used, thus providing increased confidence in our experimental setup. 
For clarity, in the text below we present our approach from the perspective of ``transfer from a source to target domain,'' with the assumption that obtaining data in the target domain is always expensive, regardless of whether it is a simulated or real-world environment. All of our experiments are conducted in Habitat \cite{habitat19iccv}.
\subsection{Task: PointGoal Navigation}
In PointNav, a robot is initialized in an unseen environment and asked to navigate to a goal location specified in relative coordinates purely from egocentric RGB-D observations without a map, in a limited time budget. An episode is considered successful if the robot issues the \texttt{STOP} command within 0.2m of the goal location. In order to increase confidence that our simulation settings will translate to the real-world, we limit episodes to 200 steps, limit number of collisions allowed (before deeming the episode a failure) to 40, and turn sliding off-- specifications found by ~\cite{srcc} to have high sim2real predictivity (how well evaluation in simulation predicts real-world performance). Sliding is a behavior enabled by default in many physics simulators that allows agents to slide along obstacles when the agent takes an action that would result in a collision. Turning sliding off ensures that the agent cannot cheat in simulation by sliding along obstacles. We use success rate (SUCC), and Success weighted by (normalized inverse) Path Length (SPL) \cite{anderson2018evaluation} as metrics for evaluation. 


\subsection{Robot in Simulation}
\vspace{-.1cm}
\xhdr{Body.} The robot has a circular base with a radius of 0.175m and a height of 0.61m. These dimensions correspond to the base width and camera height of the LoCoBot robot \cite{locobot}.

\begin{figure*}[t]
  \vspace{5 pt}
  \centering%
  \resizebox{\textwidth}{!}{
  \renewcommand{\tableTitle}[1]{\Huge{#1}}%
  \setlength{\figwidth}{\textwidth}%
  \setlength{\tabcolsep}{1.5pt}%
  \renewcommand{\arraystretch}{0.8}%
  \renewcommand{\cellset}{\renewcommand\arraystretch{0.8}%
  \setlength\extrarowheight{0pt}}%
  
  \normalsize
  \hspace{-0.25cm}\begin{tabular}{c c c c c}
  & \tableTitle{No Noise} & \tableTitle{0.1 Gaussian} & \tableTitle{0.1 Speckle} & \tableTitle{1.0 Poisson} \\
  \rotatebox{90}{\tableTitle{RGB}} &%
   \makecell{\includegraphics[width=\figwidth]{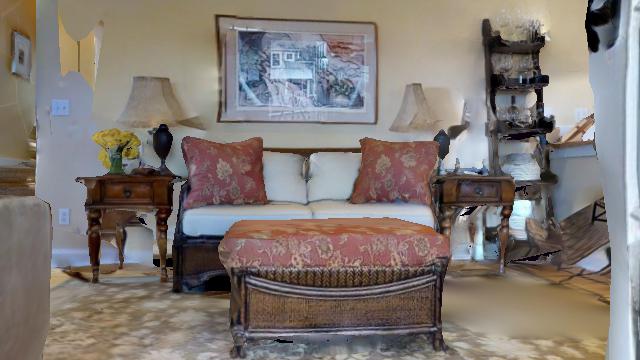}}&%
   \makecell{\includegraphics[width=\figwidth]{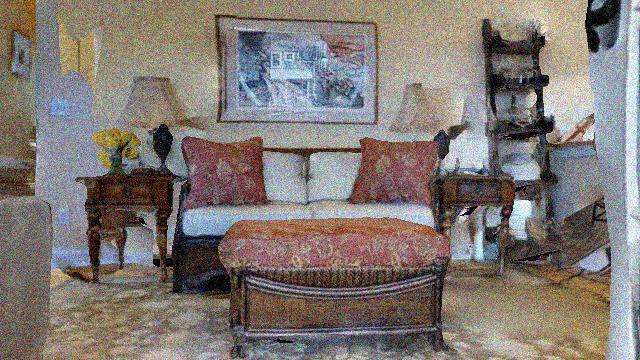}} &
   \makecell{\includegraphics[width=\figwidth]{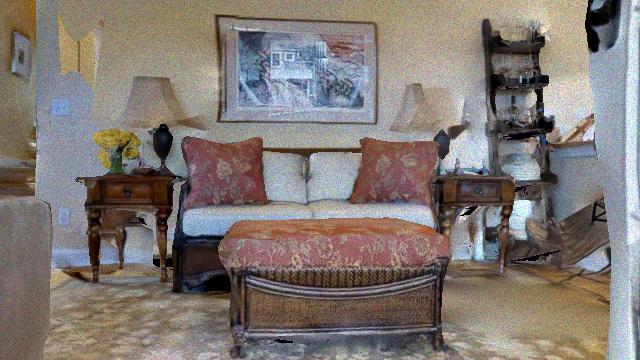}} &
   \makecell{\includegraphics[width=\figwidth]{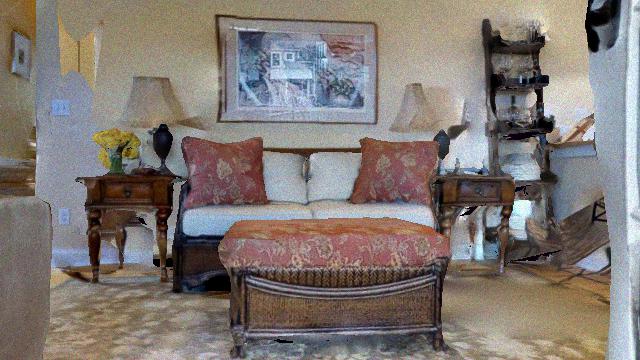}} \\\\
   & \tableTitle{No Noise} & \tableTitle{1.0 Redwood} & \tableTitle{1.5 Redwood} & \tableTitle{2.0 Redwood} \\  
   \rotatebox{90}{\tableTitle{Depth}} &%
   \makecell{\includegraphics[width=\figwidth]{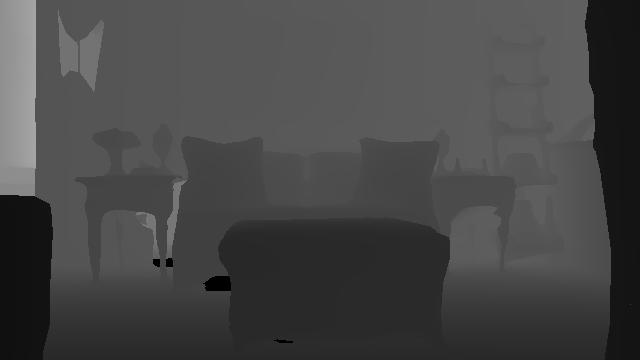}}&
  \makecell{\includegraphics[width=\figwidth]{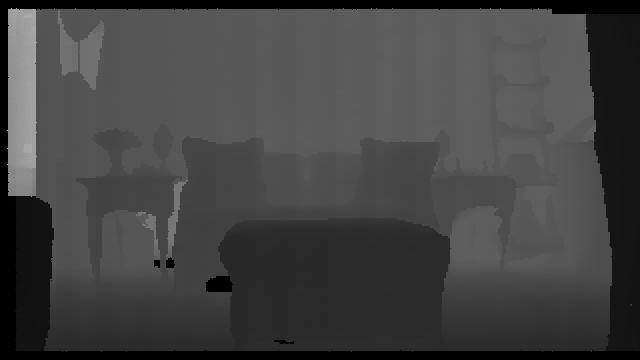}}&%
  \makecell{\includegraphics[width=\figwidth]{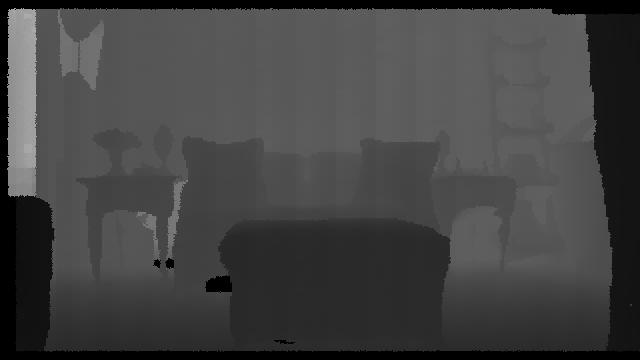}} &
  \makecell{\includegraphics[width=\figwidth]{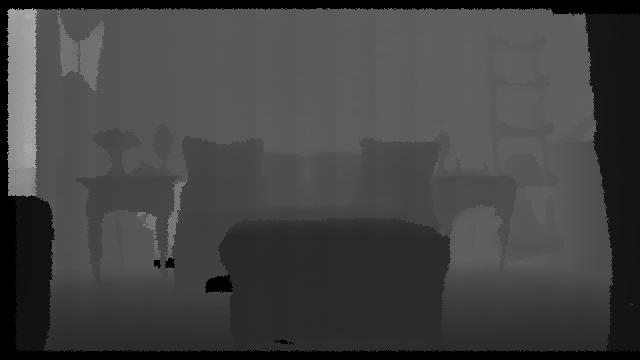}} 
  \end{tabular}}
  \caption{Different RGB and Depth sensor noise combinations we apply to our training and testing environments. From left to right: 0 RGB noise + 0 Depth noise, 0.1 Gaussian RGB noise + 1.0 Redwood Depth noise, 0.1 Speckle RGB noise + 1.5 Redwood Depth noise, 1.0 Poisson RGB noise + 2.0 Redwood Depth noise.}
  \label{fig:sensor_noise}
  \vspace{-20 pt}
  \end{figure*}

\xhdr{Sensors.} The robot has access to an egocentric RGB and Depth sensor, and accurate localization and heading through a GPS+Compass sensor. Real-world robot experiments from \cite{srcc} used Hector SLAM \cite{hector_slam} with a Hokuyo UTM-30LX LIDAR sensor and found that localization errors were approximately 7cm (much lower than the 20cm PointNav success criterion). This gives us confidence that our results will generalize to reality, despite the lack of precise localization. We match the specifications of the Intel D435 camera on the LoCoBot, and set the camera field of view to 70. To match the maximum range on the depth camera, we clip the simulated depth readings to 10m.  

\xhdr{Sensor Noise.} To simulate noisy sensor observations of the real-world, we add RGB and Depth sensor noise models to the simulator. Specifically, we use Gaussian, Speckle, and Poisson noise for the RGB camera, and Redwood noise for the Depth camera. More details about the Redwood noise can be found in ~\cite{Choi_2015_CVPR}. \reffig{fig:sensor_noise} shows a comparison between noise free RGB-D images and RGB-D images with the different noise models and multipliers we use.

\xhdr{Actions.} The action space for the robot is \texttt{turn-left 30$^{\circ}$}\, \texttt{turn-right 30$^{\circ}$}, \texttt{forward 0.25m}, and \texttt{STOP}. In the source simulator, these actions are executed deterministically and accurately. However, actions in the real-world are never deterministic -- identical actions can lead to vastly different final locations due to the actuation noise (wheel slippage, battery power drainage, etc.) typically found on a real robot. To simulate the noisy actuation that occurs in the real-world, we leverage the real-world translational and rotational actuation noise models characterized by \cite{pyrobot2019}. A Vicon motion capture was used to measure the difference in commanded state and achieved state on LoCoBot for 3 different positional controllers: Proportional Controller, Dynamic Window Approach Controller from Movebase, and Linear Quadratic Regulator (ILQR). These are controllers typically used on a mobile robot. From a state (x, y, $\theta$) and given a particular action, we add translational noise sampled from a truncated 2D Gaussian, and rotational noise from a 1D Gaussian to calculate the next state.

\subsection{Testing Environment}
\begin{figure}[ht]
  \centering%
  \vspace{-5pt}
  \resizebox{.9\columnwidth}{!}{
  \renewcommand{\tableTitle}[1]{\large{#1}}%
  \setlength{\figwidth}{0.3\columnwidth}%
  \setlength{\tabcolsep}{1.5pt}%
  \renewcommand{\arraystretch}{0.8}%
  \renewcommand{\cellset}{\renewcommand\arraystretch{0.8}%
  \setlength\extrarowheight{0pt}}%
  
  
  \hspace{-0.25cm}\begin{tabular}{c c}
   \makecell{\includegraphics[height=0.3\columnwidth]{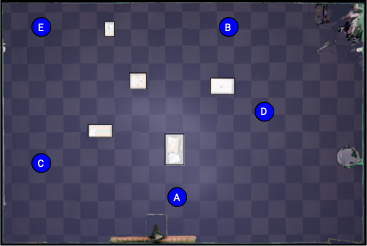}}&%
   \makecell{\includegraphics[height=0.3\columnwidth]{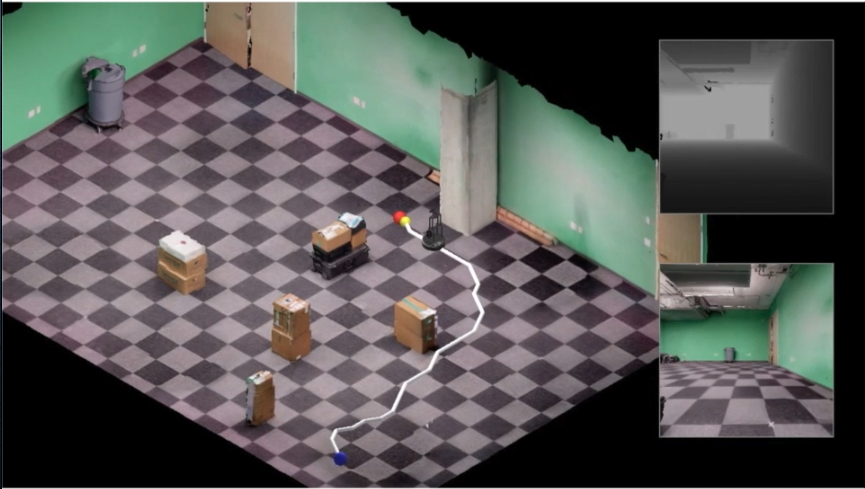}}\\
   (a) & (b)
  \end{tabular}}
  \caption{(a) Top-down view of one of our testing environments. White boxes are obstacles. The robot navigates sequentially through the waypoints $A \rightarrow B \rightarrow C \rightarrow D \rightarrow E \rightarrow A$. Figure taken from \cite{srcc}. (b) 3D visualization of the robot navigating in one of our testing environments in simulation. RGB and Depth observations are shown on the right.} 
 \vspace{-5pt}
  \label{fig:coda}
  \end{figure}

We virtualize a 6.5m by 10m real lab environment (LAB) to use as our testing environment, using a Matterport Pro2 3D camera. To model the space, we placed the Matterport camera at various locations in the room, and collected 360$^{\circ}$ scans of the environment. We used the scans to create 3D meshes of the environment, and directly imported the 3D meshes into Habitat to create a photorealistic replica of LAB \reffig{fig:coda}b. We vary the number of obstacles in LAB to create 3 room configurations with varying levels of difficulty. \reffig{fig:coda} shows one of our room configurations with 5 obstacles. We perform testing over the 3 different room configurations, each with 5 start and end waypoints for navigation episodes, and 10 independent trials, for a total of 150 runs. We report the average success rate and SPL over the 150 runs. 

Our models were trained entirely in the Gibson dataset ~\cite{xia2018gibson}, and have never seen LAB during training. The Gibson dataset  contains 3D models of $572$ cluttered indoor environments (homes, hospitals, offices, museums, etc.). In this work, we used the 72 Gibson environments that were rated 4+ in quality in ~\cite{habitat19iccv}.

\subsection{Experimental Protocol}

\begin{table}
\centering
\vspace{5 pt}

\caption{\small Definition of the 10 different noise settings we use for training and testing. Row 1 indicates the `source' environment with no observation or actuation noise present.}
\label{tab:sim_envs}
    \resizebox{.9\columnwidth}{!}{
    \begin{tabular}{cccc}
        \toprule
        \textbf{\#}  & \textbf{RGB Obs Noise} & \textbf{Depth Obs Noise} & \textbf{Actuation Noise}\\
        \midrule
        1 & - & - & -  \\
        \cmidrule(l{0pt}r{0pt}){1-4}
        2 &  Gaussian 0.1 & Redwood 1.0 \\
        3 & - & - & Proportional 1.0\\
        4  & Gaussian 0.1 & Redwood 1.0 & Proportional 1.0\\
        \cmidrule(l{0pt}r{0pt}){1-4}
        5 & Speckle 0.1 & Redwood 1.5\\
        6 & - & - & Move Base 1.0\\
        7 & Speckle 0.1 & Redwood 1.5 & Move Base 1.0\\
        \cmidrule(l{0pt}r{0pt}){1-4}
        8 & Poisson 1.0 & Redwood 2.0 & -\\ 
        9 & - & - & ILQR 1.0\\ 
        10 & Poisson 1.0 & Redwood 2.0 & ILQR 1.0\\
        \bottomrule
    \end{tabular}}
    \vspace{-15pt}
\end{table}

Recall that our objective is to improve the ability for RL agents to generalize to new environments using little real-world data. To do this, we define our source environment as Gibson without any sensor or actuation noise (\gibnn). We create 10 target environments with noise settings described in \reftab{tab:sim_envs}. We use the notation $O$ to represent an environment afflicted with only RGBD observation noise (rows 2, 5, or 8), $D$ to represent an environment afflicted with only dynamics noise (rows 3, 6, or 9), and $O+D$ to represent an environment afflicted with RGBD observation noise and dynamics noise (rows 4, 7, or 10). 

\subsection{RL Navigation Models}
\label{sec:nav_models}
We train learning-based navigation policies, $\pi$, for PointNav in Habitat using environments from the Gibson dataset. Policies were trained from scratch with reinforcement learning using DD-PPO ~\cite{ddppo}, a decentralized, distributed variant of the proximal policy optimization (PPO) algorithm, that allows for large-scale training in GPU-intensive simulation environments. We follow the navigation policy described in ~\cite{ddppo}, which is composed of a ResNet50 visual encoder, and a 2-layer LSTM. Each policy is trained using 64 Tesla V100s. Base policies are trained for 100 million steps ($\pi^{100M}$) to ensure convergence.

%% file: sections/5_experiments.tex
\section{Experiments}
Our experiments aim to answer the following: (1) How large is the sim2real gap? 
(2) Does our method improve generalization to target domains? (3) How does our method compare to directly training (or fine-tuning) in the target environment? (4) How much real-world data do we need? 

  
  

\smallskip
\xhdr{How large is the sim2real gap?}
\label{sec:results_1}
First, we show that RL policies fail to generalize to new environments. We train a policy without any noise (\pis), and a policy with observation and dynamics noise (\pit). 
We test these policies in LAB with 4 different noise settings: \labnn, \labsens, \labact, \labn, and average across the noise settings. For each noise setting, we conduct 3 sets of runs, each containing 150 episodes in the target environments. We see that \pis tested in \labnn exhibits good transfer across environments -- 0.84 SPL (in contrast, the Habitat 2019 challenge winner was at 0.95 SPL ~\cite{habitat_challenge}). ~\cite{ddppo} showed that near-perfect performance is possible when the policy is trained out for significantly longer (2.5B frames), but for the sake of multiple experiments, we limit our analysis to 100M frames of training and compare all models across the same number.

\begin{figure}[h]
  \centering
  \vspace{-5 pt}
  \includegraphics[width=.9\columnwidth]{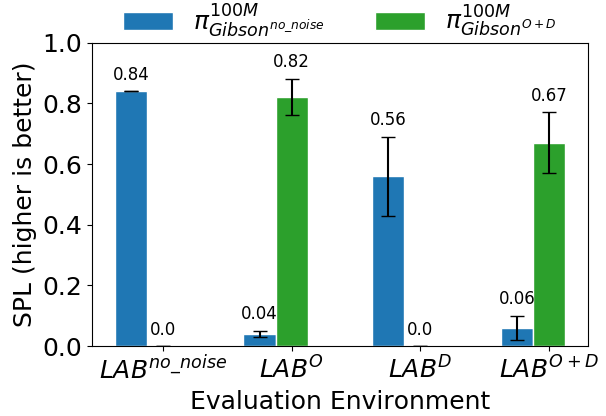}
  \caption{Zero-shot transfer of \pis and \pit tested in LAB with different combinations of observation and dynamics noise. We see that SPL drops when a policy is tested in an environment with noise different from what it was trained in.}
  \label{fig:pi_st_avg_zeroshot}
  \vspace{-10pt}
\end{figure}

From \reffig{fig:pi_st_avg_zeroshot}, we see that when dynamics noise is introduced (\pis tested in \labact), SPL drops from 0.84 to 0.56 (relative drop of 28\%). More significantly, when sensor noise is introduced (\pis tested in \labsens), SPL drops to 0.04 (relative drop of 81\%), and when both sensor and dynamics noise are present, (\pis tested in \labn), SPL drops to 0.06 (relative drop of 78\%). Thus, in the absence of noise, generalization across scenes (Gibson to LAB) is good, but in the presence of noise, the generalization suffers. We also notice that the converse is true: policies trained from scratch in \gibn environments fail to generalize to \labnn and \labact environments. These results show us that RL agents are highly sensitive to what might be considered perceptually minor changes to visual inputs. To the best of our knowledge, no prior work in embodied navigation appears to have considered this question of sensitivity to noise; hopefully our results will encourage others to consider this as well. 

\smallskip
\xhdr{How well does \oa do?}
\label{sec:oa}
\begin{table*}[t!]
    \vspace{10 pt}
    \caption{Success rate and SPL of five policies with RGB-D observations. \pis is a policy trained solely in simulation. \piroa is \pis equipped with \oa trained using 5k images from the source and target environments. \pirda, \pir and \pigt are initialized with \pis. \pirda is fine-tuned with \da using 5k samples from the target environment. \pir is finetuned using the full BDA pipeline, utilizing both \oa and \da. \pigt is fine-tuned directly in the target environment for 1M steps of experience, and serves as an oracle baseline. While \pigt and \pir achieve in strong performance across environments with varying noises (rows 4, 8, 12), BDA requires 200$\times$ fewer samples from the target environment.}
    \resizebox{\textwidth}{!}{
    \begin{tabular}{c c c c cc cc cc cc cc}
    \toprule
    \texttt{\#} & {\bf RGB Obs} & \bf{Depth Obs} & {\bf Actuation} & \multicolumn{2}{c}{\pis} & \multicolumn{2}{c}{\piroa} & \multicolumn{2}{c}{\pirda} & \multicolumn{2}{c}{\pir} & \multicolumn{2}{c}{\pigt}\\
    \cmidrule(l{4pt}r{4pt}){5-6}
    \cmidrule(l{4pt}r{0pt}){7-8}
    \cmidrule(l{4pt}r{0pt}){9-10}
    \cmidrule(l{4pt}r{0pt}){11-12}
    \cmidrule(l{4pt}r{0pt}){13-14}
    & \bf{Noise} & \bf{Noise} & \bf{Noise} & SUCC & SPL & SUCC & SPL & SUCC & SPL & SUCC & SPL & SUCC & SPL\\
    \midrule
    1  & - & - & - & 1.00 & 0.84 & 1.00 & 0.85 & 1.00 & 0.89 & 0.80 & 0.61 & 0.99 & 0.84 \\
    2 & Gaus. 0.1 & Red. 1.0 & - & 0.10 & 0.04 & 1.00 & 0.78 & 0.21 & 0.10 & 0.97 & 0.80 & 0.99 & 0.87  \\
    3 & - & - & Prop. 1.0 & 0.89 & 0.57 & 0.86 & 0.54 & 1.00 & 0.66 & 0.99 & 0.65 & 1.00 & 0.64 \\
    4 & Gaus. 0.1 & Red. 1.0 & Prop. 1.0 & 0.32 & 0.11 & 0.78 & 0.48 & 0.16 & 0.05 & 1.00 & 0.62 & 1.00 & 0.65 \\
    \cmidrule(l{0pt}r{0pt}){2-14}
    
    5 & - & - & - & 1.00 & 0.84 & 1.00 & 0.85 & 0.98 & 0.80 & 0.85 & 0.68 & 0.97 & 0.80  \\ 
    6 & Speck. 0.1 & Red. 1.5 & - & 0.11 & 0.05 & 1.00 & 0.80 & 0.03 & 0.01 & 0.70 & 0.54 & 0.99 & 0.81  \\
    7 & - & - & MB 1.0 & 0.71 & 0.42 & 0.79 & 0.47 & 1.00 & 0.59 &  0.97 & 0.58 & 1.00 & 0.59 \\
    8 & Speck. 0.1 & Red. 1.5 & MB 1.0 & 0.08 & 0.03 & 0.68 & 0.39 & 0.03 & 0.01 & 0.99 & 0.60 & 1.00 & 0.62 \\
    
    \cmidrule(l{0pt}r{0pt}){2-14}
    9 & - & - & -  & 1.00 & 0.85 & 1.00 & 0.86 & 1.00 & 0.85 & 1.00 & 0.87 & 1.00 & 0.83 \\
    10 & Pois. 1.0 & Red. 2.0 & - & 0.07 & 0.04 & 0.68 & 0.51 & 0.25 & 0.13 & 0.96 & 0.69 & 1.00 & 0.87 \\
    11 & - & - & ILQR 1.0 & 0.93 & 0.68 & 0.95 & 0.69 & 1.00 & 0.74 & 1.00 & 0.76 & 0.99 & 0.73 \\
    12 & Pois. 1.0 & Red. 2.0 & ILQR 1.0 & 0.14 & 0.05 & 0.63 & 0.39 & 0.25 & 0.08 & 0.99 & 0.63 & 1.00 & 0.73\\
    \bottomrule
    \end{tabular}
    }
    \label{tab:results}
    \vspace{-15pt}
\end{table*}
\setlength{\tabcolsep}{1.4pt}
Following Alg. \ref{alg:architecture} described in Sec. \ref{sec:domain_adapt}, we train \oa from scratch for 200 epochs. In \reffig{fig:cyc}, we see that the model learns to remove the Gaussian noise placed on the RGB image, and learns to smooth out textures in the depth image. In Table \ref{tab:results}, we see that simply equipping \pis with \oa during deployment (\piroa) drastically improves SPL in \labsens compared to \pis, resulting in an average increase of 65\% (rows 2, 6, 10).

\vspace{-5pt}
\begin{figure}[h]
  \centering%
  \resizebox{.95\columnwidth}{!}{
  \renewcommand{\tableTitle}[1]{\large{#1}}%
  \setlength{\figwidth}{0.3\columnwidth}%
  \setlength{\tabcolsep}{1.5pt}%
  \renewcommand{\arraystretch}{0.8}%
  \renewcommand\cellset{\renewcommand\arraystretch{0.8}%
  \setlength\extrarowheight{0pt}}%
  
  
  \hspace{-0.25cm}\begin{tabular}{c c c c}
   & \imgnn & \imgon & \oa(\imgon)\\
   \rotatebox[origin=c]{90}{RGB} &%
   \makecell{\includegraphics[width=\figwidth]{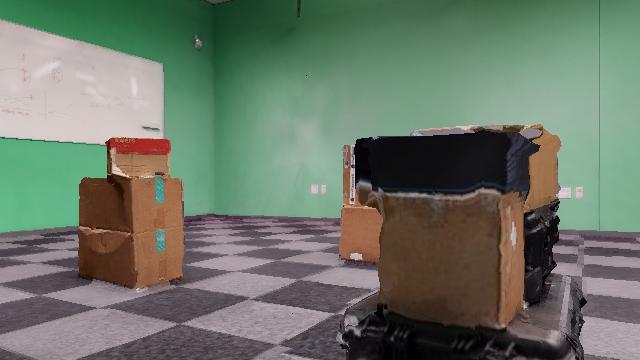}} &%
   \makecell{\includegraphics[width=\figwidth]{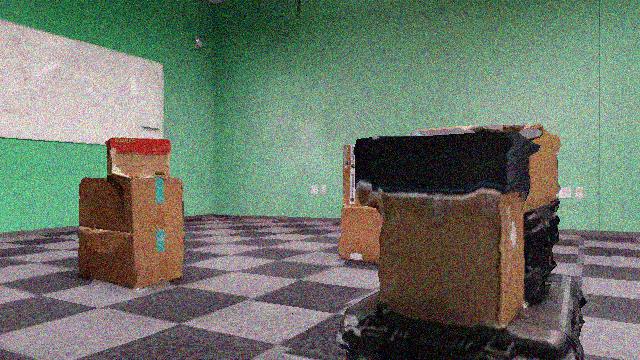}} &
   \makecell{\includegraphics[width=\figwidth]{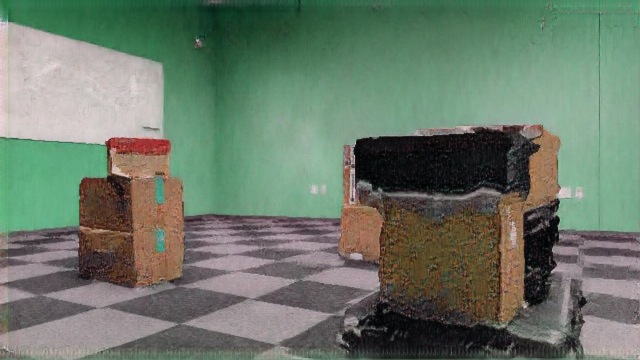}} \\
   \rotatebox[origin=c]{90}{Depth} &%
    \makecell{\includegraphics[width=\figwidth]{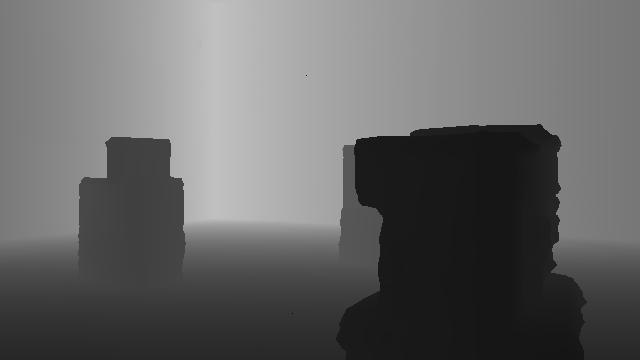}} &
   \makecell{\includegraphics[width=\figwidth]{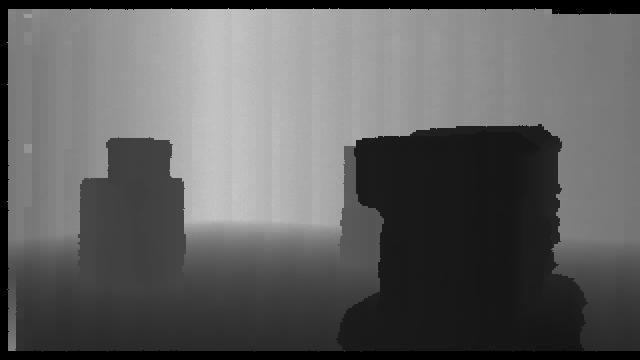}} &%
   \makecell{\includegraphics[width=\figwidth]{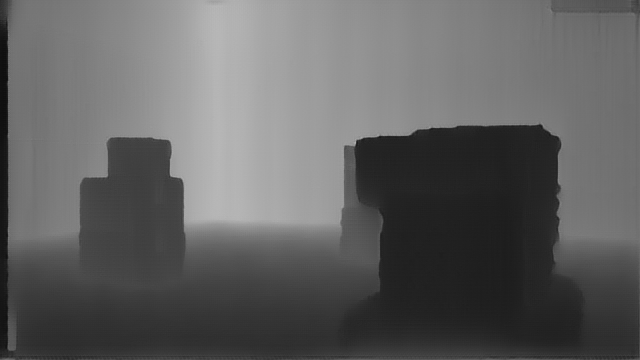}} \\
   & (a) & (b) & (c)
  \end{tabular}}
  \caption{(a) LAB with no sensor noise (b) LAB with 0.1 Gaussian RGB noise and 1.0 Redwood Depth noise. (c) By adapting images from \emph{real2sim}, we now have images that closely resemble (a).}
  \label{fig:cyc}
  \vspace{-5pt}
  \end{figure}

In addition, we have RGB-D images of LAB collected from a real robot, pre-COVID, and results using our real2sim \oa module. While no GAN metric is perfect (user studies are typically conducted for evaluation as done in ~\cite{CycleGAN2017}), we calculated the Fréchet Inception Distance (FID) ~\cite{heusel2017gans} score (lower is better) to provide quantitative results. We find that the FID comparing \ireal and \isim is 100.74, and the FID from \oa(\ireal) to \isim images is 83.05. We also calculated FID comparing simulation images afflicted with Gaussian noise, \imggaus, to noise-free simulation images \imgnn to be 98.73, and FID between \oa(\imggaus) to \imgnn images to be 88.44. To put things in context, the FID score comparing images from CIFAR10 to our simulation images is 317.61. This shows that perceptually, the distribution of our adapted images more closely resembles images taken directly from simulation, and that real2sim \oa is not far off from our sim2sim \oa experiments. While our architecture has changed since this initial data collection (initial images are 256 $\times$ 256, compared to our current architecture which uses 640 $\times$ 360 images), these results will serve as a good indication that our approach will generalize to reality.

\begin{figure}[h]
  \vspace{-5 pt}
  \centering%
  \resizebox{0.8\columnwidth}{!}{
  \renewcommand{\tableTitle}[1]{\small{#1}}%
  \setlength{\figwidth}{0.3\columnwidth}%
  \setlength{\tabcolsep}{1.5pt}%
  \renewcommand{\arraystretch}{0.6}%
  \renewcommand\cellset{\renewcommand\arraystretch{0.6}%
  \setlength\extrarowheight{0pt}}%
  
  \scriptsize
  \hspace{-0.25cm}\begin{tabular}{c c c}
   &\tableTitle{\ireal}&\tableTitle{\oa(\ireal)}\\
   \rotatebox[origin=c]{90}{\tableTitle{RGB}}&%
   \makecell{\includegraphics[width=\figwidth]{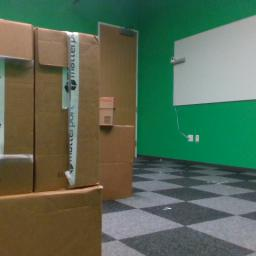}} &
   \makecell{\includegraphics[width=\figwidth]{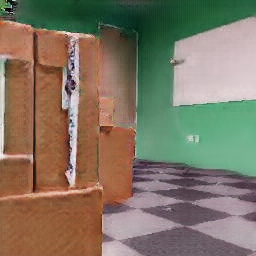}}
    \\
   \rotatebox[origin=c]{90}{\tableTitle{Depth}}&%
   
   \makecell{\includegraphics[width=\figwidth]{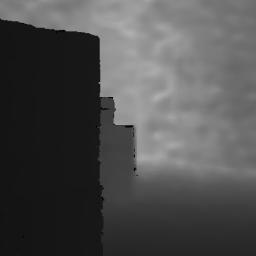}} &
   \makecell{\includegraphics[width=\figwidth]{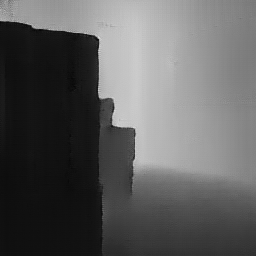}} \\
   
   & (a) & (b) 
  \end{tabular}}
  \caption{(a) LAB$^{\text{real}}$ (b) We adapt from \emph{real2sim} to obtain images that closely resemble RGB-D images from simulation.}
  \label{fig:simvreal}
  \vspace{-15pt}
  \end{figure}

\begin{table}[h]
\centering
\vspace{5 pt}
\caption{\small Average translation and rotation actuation error for LoCoBot using the PyRobot proportional controller. For a given action, the actuation error is sampled from the noise models, and added to the action to calculate the next state. We report the noise model characterized by real-world benchmarking by PyRobot, as well as the learned \da noise model from real-world experiments in LAB.}
\label{tab:da}
    \begin{tabular}{l c c c}
        \toprule
        \textbf{}  & \textbf{X error (mm)} & \textbf{Y error (mm)} & \textbf{$\theta$ error (rad)}\\
        \midrule
        \textbf{PyRobot} & & &  \\
        \quad \textbf{Linear motion} & 0.042 $\pm{0.15}$ & 0.017 $\pm{0.08}$ & 0.031 $\pm{0.16}$\\
        \quad \textbf{Rotation motion} & 0.005 $\pm{0.06}$ & 0.001 $\pm{0.03}$ & 0.043 $\pm{0.13}$ \\
        \cmidrule(l{0pt}r{0pt}){1-4}
        \textbf{LAB} & & &  \\
        \quad \textbf{Linear motion} & 0.093 $\pm{0.08}$ & 0.016 $\pm{0.15}$ & 0.002 $\pm{0.00}$ \\
        \quad \textbf{Rotation motion} & 0.001 $\pm{0.00}$ & 0.012 $\pm{0.01}$ & 0.004 $\pm{0.01}$ \\
        \bottomrule
    \end{tabular}
    \vspace{-20pt}
\end{table}

\smallskip
\xhdr{How well does \da do?}
\label{sec:da}
We train \da using 5,000 samples of state-action pairs collected in the target environment, and augment our source simulator with \da. From Table \ref{tab:results}, we see that finetuning \pis with \da (\pirda) on average, leads to a relative 15\% improvement in success and a 11\% improvement in SPL over \pis in \labact (rows 3, 7, 11). 
Next, we investigate how well our actuation noise model approximates real-world conditions. Using state-action pairs collected from LoCoBot in LAB using the PyRobot proportional controller from our experiments pre-COVID, we trained a \da module to approximate the translation and rotation noise present in our real-world testing environment. We compare this to the actuation noise models used in our target environments, which were provided from the real-world benchmark by PyRobot ~\cite{pyrobot2019} using a Vicon motion capture system. Since actuation noise models are a factor of the robot and environment, the \da learned using our real-world experiments cannot exactly match the noise model benchmarked by PyRobot. However, Table \ref{tab:da} shows that the noise model learned from LAB is similar in order of magnitude to the noise models derived PyRobot. This gives us confidence that the actuation noise models used in our target simulation as a stand in for reality are a good approximation for the dynamics noise present in the real-world.

\begin{figure*}[t!]
  \centering%
  \vspace{5pt}
  \resizebox{\textwidth}{!}{
  \renewcommand{\tableTitle}[1]{\large{#1}}%
  \setlength{\figwidth}{0.3\textwidth}%
  \setlength{\tabcolsep}{1.5pt}%
  \renewcommand{\arraystretch}{0.8}%
  \renewcommand\cellset{\renewcommand\arraystretch{0.8}%
  \setlength\extrarowheight{0pt}}%
  
  
  \hspace{-0.25cm}\begin{tabular}{c c c}
   \makecell{\includegraphics[width=0.3\columnwidth]{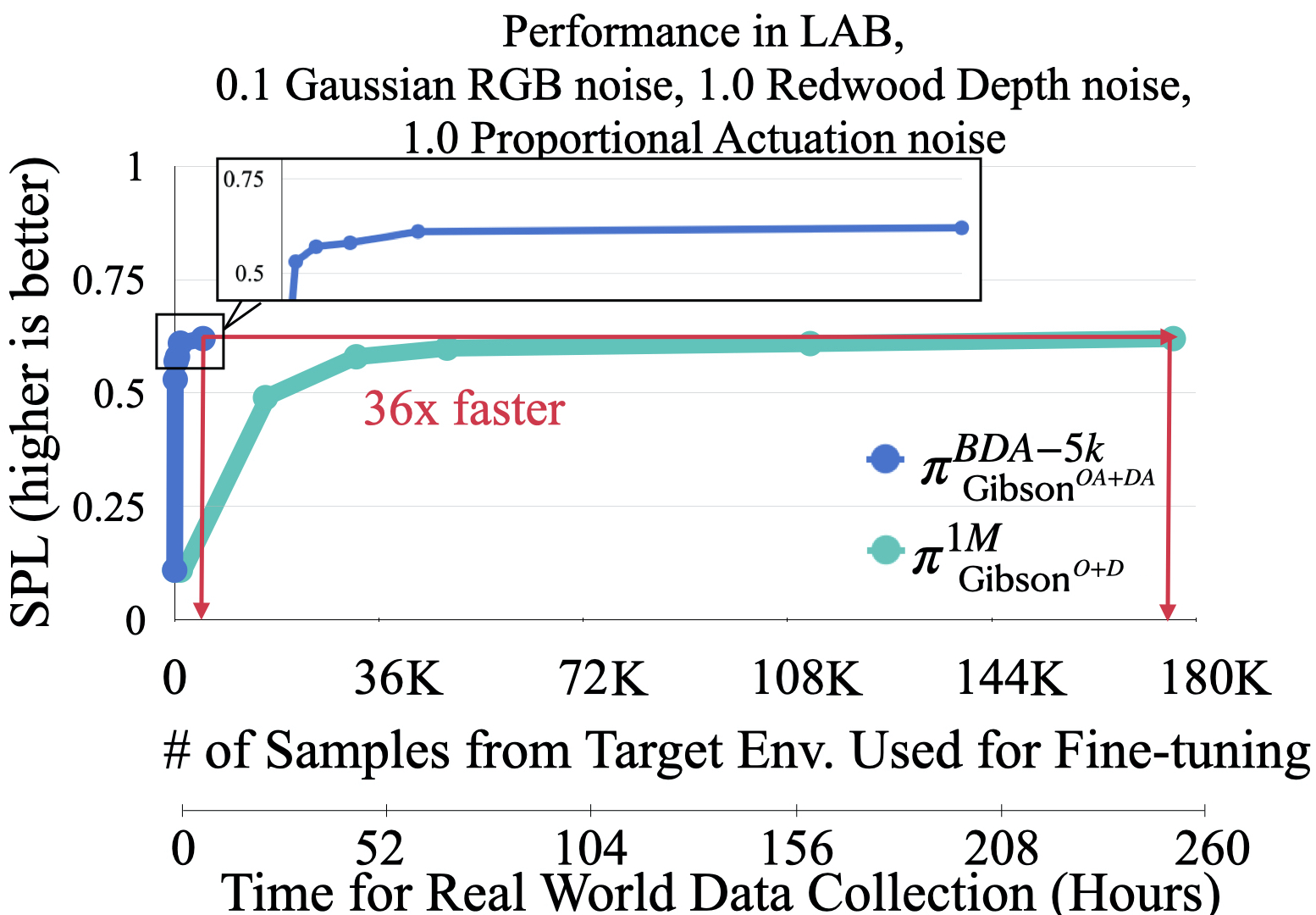} }&%
   \makecell{\includegraphics[width=0.3\columnwidth]{figures/sample_efficiency/s2.png}} &
   \makecell{\includegraphics[width=0.3\columnwidth]{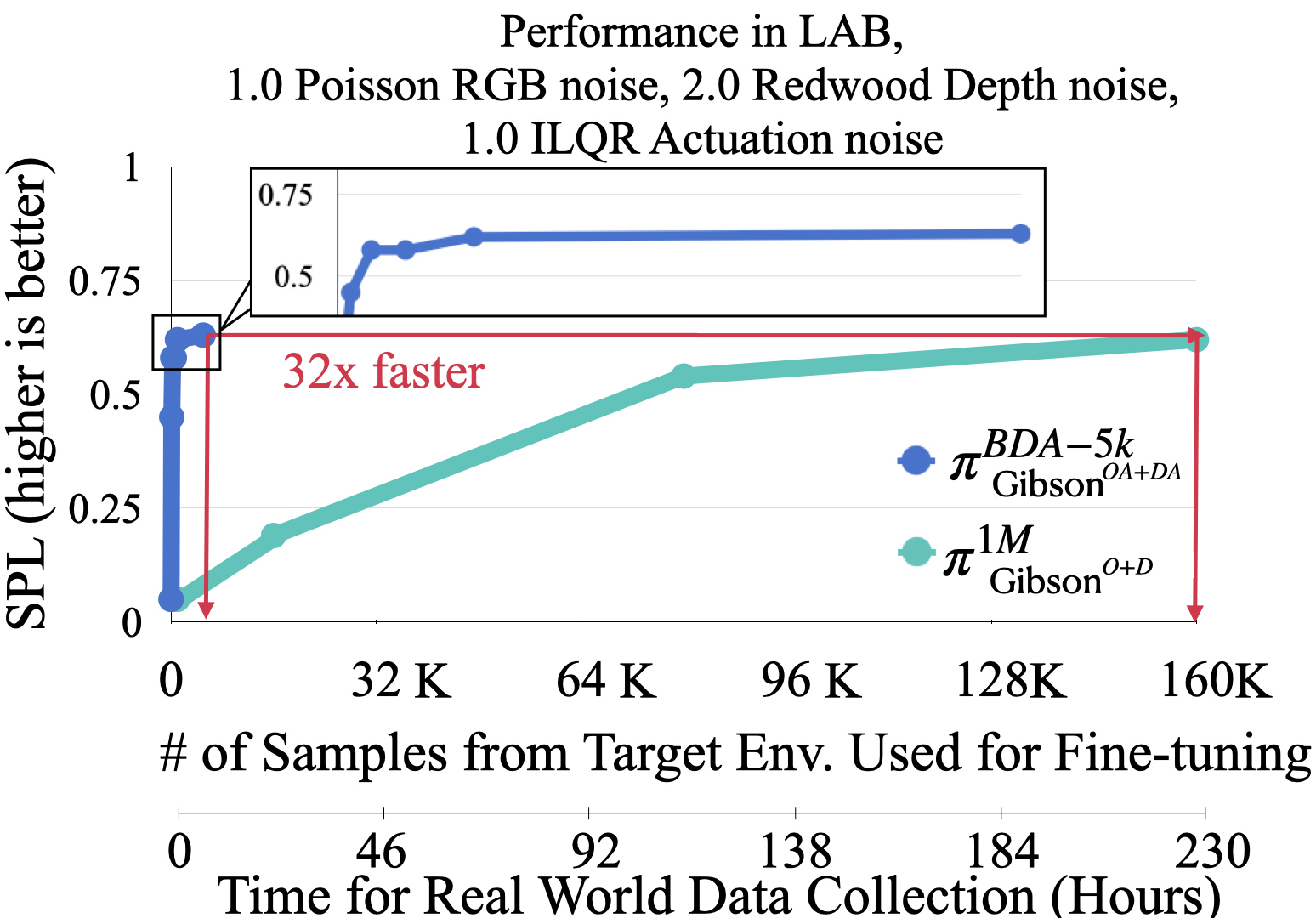}} \\
   \tiny{(a)} & \tiny{(b)} & \tiny{(c)} 
  \end{tabular}}
  \caption{Performance of BDA compared to directly finetuning a policy in the target environment: Plots (a), (b) and (c) represent LAB environments with different noise settings we test in. On average, BDA requires 61$\times$ less data from the target environment to achieve the same SPL as finetuning directly in the target environment.}
  \label{fig:sample_efficiency_zoom}
  \vspace{-20pt}
  \end{figure*}

\smallskip
\xhdr{How does our method compare to fine-tuning?}
\label{sec:results_2_3}
We evaluate our policy finetuned using BDA with 5,000 data samples collected in the target environment (\pir). We compare this to directly finetuning in the target environment (\pigt), which serves as an oracle baseline. Both \pir and \pigt are initialized with \pis, and both are re-trained for each target $O+D$ setting.

Our results in \reftab{tab:results} show the benefits in finetuning with BDA using data from target environments. While \piroa and \pirda show improvements over the baseline policy, both policies still fail to generalize to environments in which new noise is present. Specifically, \piroa fails to generalize to \labact and \labn environments, and \pirda fails to generalize to \labsens and \labn environments. On the other hand, both \pir and \pigt demonstrate robustness in all combinations of sensor and actuation noise. We also observe that using BDA to learn the observation and dynamics noise models with 5,000 samples from the target environment is capable of nearly matching performance of \pigt. In fact, we only see, on average, a 5\% difference between \pigt and \pir (rows 4, 8, 12), while the former is directly trained in the target environment which is not possible in reality, as it requires 1M samples from the target environment. In contrast, we see on average, a 25\% difference between \pigt and \piroa, and an average 62\% difference between \pigt and \pirda (rows 4, 8, 12). This highlights the importance of our proposed framework to close the reality gap in both directions; to utilize both real2sim observation adaptation and sim2real dynamics adaptation to accommodate for variations that are overlooked by approaches that only focus on one of these two directions.

From these results, we notice in certain environments our method performs worse than the oracle baseline if no or only observation noise is present (rows 1, 5, 6, 10), but performs on the level of the oracle baseline when dynamics is added (rows 3, 4, 7, 8, 11, 12). We believe it’s due to `sliding’, a default behavior in 3D simulators allowing agents to slide along obstacles rather than stopping immediately on contact. Following the findings and recommendations of ~\cite{srcc}, we disabled sliding to make our simulation results more predictive of real-world experiments. We find that one common failure mode in the absence of sliding is that agents get stuck on obstacles. In the presence of dynamics noise, the slight amount of actuation noise allows the agent to free itself from obstacles, similar to how it would in reality. Without dynamics noise, the agents continue to stay stuck. 



\smallskip
\xhdr{Sample Efficiency.}
We repeat our experiments, varying the amounts of data collected from the target environment. We re-train \oa and \da using 100, 250, 500, 1,000, and 5,000 steps of experience in the target environment, and re-evaluate performance. We compare this to directly finetuning in the target environment for varying amounts of data. 

In \reffig{fig:sample_efficiency_zoom}, the x-axis represents the number of samples collected in the target environment. From previous experiments, we estimate 1 episode in the real-world to last on average 6 minutes, in which the robot will take approximately 70 steps to reach the goal. We use this as a conversion factor, and add an additional x-axis to show the number of hours needed for collecting the required samples from the target environment. The y-axis shows the SPL in the target environment. 
We see that the majority of our success comes from our first 1,000 samples from the target environment, and after 5,000 samples, \pir is able to match the performance from \pigt. Collecting 5,000 samples of data from a target environment to train our method would have taken 7 hours. In comparison, \reffig{fig:sample_efficiency_zoom}b shows that we would have to finetune the base policy for approximately 585,000 steps in the target environment (836 hours to collect data from target environment) to reach the same SPL. Comparing the amount of data needed to reach the same SPL, we see that BDA reduces the amount of data needed from the target environment by 36$\times$ in \reffig{fig:sample_efficiency_zoom}a, 117$\times$ in \reffig{fig:sample_efficiency_zoom}b, and 32$\times$ in \reffig{fig:sample_efficiency_zoom}c, for an average speed up of 61$\times$.
These results give us confidence in the importance of our approach, as we wish to limit the amount of data needed from a target environment (i.e. real-world). 

%% file: sections/6_related_work.tex
\section{Related Work}
\label{sec:citations}
\ouralgfull is related to literature on domain and dynamics randomization, domain adaptation, and residual policy learning. 

\smallskip
\xhdr{Domain and Dynamics Randomization.} Borrowing ideas from data augmentation commonly used in computer vision, domain randomization is a technique to train robust policies by exposing the agent to a wide variety of simulation environments with randomized visual properties such as lighting, texture, camera position, etc. Similarly, dynamics randomization is a process that randomizes physical properties in the simulator such as friction, mass, damping, etc. \cite{DBLP:SadeghiL17} applied randomization to textures to learn real indoor flight by training solely in simulation. \cite{simopt} used real-world roll outs to learn a distribution of simulation dynamics parameters to randomize over. \cite{andrychowicz2020learning} randomized both physical and visual parameters to train a robotic hand to perform in hand manipulation. However, finding the right distribution to randomize parameters over is difficult, and requires expert knowledge. If the distribution chosen to randomize parameters over is too large, the task becomes much harder for the policy to learn; if the distribution is too small, then the reality gap remains large, and the policy will fail to generalize.

\smallskip
\xhdr{Domain Adaptation.} To bridge the simulation to reality gap, many works have used domain adaptation, a technique in which data from a source domain is adapted to more closely resemble data from a target domain. Prior works have used domain adaptation techniques for adapting vision-based models to translate images from sim-to-real during training for manipulation tasks ~\cite{graspgan, rcann}, and real-to-sim during testing for navigation tasks \cite{vrgoogles}. Other works have focused on adapting policies for dynamic changes \cite{golemo2018sim, yu2020learning}. In our work, we seek to use domain adaptation to close the gap for both the visual and the dynamics domain. 

\smallskip
\xhdr{Residual Policy Learning.} An alternative to typical transfer learning techniques is to directly improve the underlying policy itself. Instead of re-training an agent from scratch when policies perform sub-optimally, the sub-optimal policy can be used as prior knowledge in RL to speed up training. This is the main idea behind residual policy learning, in which a residual policy is used to augment an initial policy to correct for changes in the environment. \cite{residualRL, Silver2018ResidualPL} demonstrated that combining residual policy learning with conventional robotic control improves the robot's ability to adapt to variations in the environment for manipulation tasks. Our method builds on this line of research by augmenting the simulator using a neural network that learns the residual error between simulation and reality.

%% file: sections/7_conclusion.tex
\section{Conclusion}
\label{sec:conclusion}
We introduce Bi-directional Domain Adaptation (BDA), a method to utilize the differences between simulation and reality to accelerate learning and improve generalization of RL policies. We use domain adaptation techniques to transfer images from \emph{real2sim} to close the visual domain gap, and learn the residual error in dynamics from \emph{sim2real} to close the dynamics domain gap. We find that our method consistently improves performance of the initial policy $\pi$ while remaining sample efficient. 

%% file: sections/8_acknowledgements.tex
\vspace{-0.5mm}
\section{Acknowledgements}
\footnotesize{The Georgia Tech effort was supported in part by NSF, AFRL, DARPA, ONR YIPs, ARO PECASE. JT was supported by an NSF Graduate Research Fellowship under Grant No. DGE-1650044 and a Google Women Techmaker’s Fellowship. The views and conclusions contained herein are those of the authors and should not be interpreted as necessarily representing the official policies or endorsements, either expressed or implied, of the U.S. Government, or any sponsor.

\noindent \textbf{License for dataset used} Gibson Database of Spaces. License at \url{http://svl.stanford.edu/gibson2/assets/GDS_agreement.pdf}}

%% file: root.bbl
\begin{thebibliography}{10}
\providecommand{\url}[1]{#1}
\csname url@rmstyle\endcsname
\providecommand{\newblock}{\relax}
\providecommand{\bibinfo}[2]{#2}
\providecommand\BIBentrySTDinterwordspacing{\spaceskip=0pt\relax}
\providecommand\BIBentryALTinterwordstretchfactor{4}
\providecommand\BIBentryALTinterwordspacing{\spaceskip=\fontdimen2\font plus
\BIBentryALTinterwordstretchfactor\fontdimen3\font minus
  \fontdimen4\font\relax}
\providecommand\BIBforeignlanguage[2]{{%
\expandafter\ifx\csname l@#1\endcsname\relax
\typeout{** WARNING: IEEEtran.bst: No hyphenation pattern has been}%
\typeout{** loaded for the language `#1'. Using the pattern for}%
\typeout{** the default language instead.}%
\else
\language=\csname l@#1\endcsname
\fi
#2}}

\bibitem{ddppo}
E.~Wijmans, A.~Kadian, A.~Morcos, S.~Lee, I.~Essa, \emph{et~al.}, ``{DD-PPO}:
  Learning near-perfect pointgoal navigators from 2.5 billion frames,'' in
  \emph{ICLR}, 2020.

\bibitem{andrychowicz2020learning}
O.~M. Andrychowicz, B.~Baker, M.~Chociej, R.~Jozefowicz, B.~McGrew,
  \emph{et~al.}, ``Learning dexterous in-hand manipulation,'' \emph{The
  International Journal of Robotics Research}, vol.~39, no.~1, pp. 3--20, 2020.

\bibitem{haarnoja2018learning}
T.~Haarnoja, S.~Ha, A.~Zhou, J.~Tan, G.~Tucker, and S.~Levine, ``Learning to
  walk via deep reinforcement learning,'' \emph{arXiv preprint
  arXiv:1812.11103}, 2018.

\bibitem{graspgan}
K.~Bousmalis, A.~Irpan, P.~Wohlhart, Y.~Bai, M.~Kelcey, \emph{et~al.}, ``Using
  simulation and domain adaptation to improve efficiency of deep robotic
  grasping,'' in \emph{2018 IEEE international conference on robotics and
  automation (ICRA)}.\hskip 1em plus 0.5em minus 0.4em\relax IEEE, 2018, pp.
  4243--4250.

\bibitem{golemo2018sim}
F.~Golemo, A.~A. Taiga, A.~Courville, and P.-Y. Oudeyer, ``Sim-to-real transfer
  with neural-augmented robot simulation,'' in \emph{Conference on Robot
  Learning}, 2018, pp. 817--828.

\bibitem{rcann}
S.~James, P.~Wohlhart, M.~Kalakrishnan, D.~Kalashnikov, A.~Irpan,
  \emph{et~al.}, ``Sim-to-real via sim-to-sim: Data-efficient robotic grasping
  via randomized-to-canonical adaptation networks,'' in \emph{Proceedings of
  the IEEE Conference on Computer Vision and Pattern Recognition}, 2019.

\bibitem{peng2018sim}
X.~B. Peng, M.~Andrychowicz, W.~Zaremba, and P.~Abbeel, ``Sim-to-real transfer
  of robotic control with dynamics randomization,'' in \emph{2018 IEEE
  international conference on robotics and automation (ICRA)}.

\bibitem{tobin2017domain}
J.~Tobin, R.~Fong, A.~Ray, J.~Schneider, W.~Zaremba, and P.~Abbeel, ``Domain
  randomization for transferring deep neural networks from simulation to the
  real world,'' in \emph{2017 IEEE/RSJ International Conference on Intelligent
  Robots and Systems (IROS)}.\hskip 1em plus 0.5em minus 0.4em\relax IEEE,
  2017.

\bibitem{yu2020learning}
W.~Yu, J.~Tan, Y.~Bai, E.~Coumans, and S.~Ha, ``Learning fast adaptation with
  meta strategy optimization,'' \emph{IEEE Robotics and Automation Letters},
  vol.~5, no.~2, pp. 2950--2957, 2020.

\bibitem{vrgoogles}
J.~Zhang, L.~Tai, P.~Yun, Y.~Xiong, M.~Liu, \emph{et~al.}, ``Vr-goggles for
  robots: Real-to-sim domain adaptation for visual control,'' \emph{IEEE
  Robotics and Automation Letters}, vol.~4, no.~2, pp. 1148--1155, 2019.

\bibitem{xia2018gibson}
F.~Xia, A.~R. Zamir, Z.~He, A.~Sax, J.~Malik, and S.~Savarese, ``Gibson env:
  Real-world perception for embodied agents,'' in \emph{CVPR}, 2018.

\bibitem{CycleGAN2017}
J.-Y. Zhu, T.~Park, P.~Isola, and A.~A. Efros, ``Unpaired image-to-image
  translation using cycle-consistent adversarial networks,'' in \emph{Computer
  Vision (ICCV), 2017 IEEE International Conference on}, 2017.

\bibitem{schapire_ijcai99}
R.~E. Schapire, ``A brief introduction to boosting,'' in \emph{Proceedings of
  the 16th International Joint Conference on Artificial Intelligence - Volume
  2}, ser. IJCAI’99, 1999.

\bibitem{anderson2018evaluation}
P.~Anderson, A.~Chang, D.~S. Chaplot, A.~Dosovitskiy, S.~Gupta, \emph{et~al.},
  ``{On Evaluation of Embodied Navigation Agents},'' \emph{arXiv preprint
  arXiv:1807.06757}, 2018.

\bibitem{habitat19iccv}
M.~Savva, A.~Kadian, O.~Maksymets, Y.~Zhao, E.~Wijmans, \emph{et~al.},
  ``Habitat: {A} {P}latform for {E}mbodied {AI} {R}esearch,'' in \emph{ICCV},
  2019.

\bibitem{srcc}
A.~Kadian, J.~Truong, A.~Gokaslan, A.~Clegg, E.~Wijmans, \emph{et~al.},
  ``Sim2real predictivity: Does evaluation in simulation predict real-world
  performance,'' \emph{IEEE Robotics and Automation Letters}, 2020.

\bibitem{locobot}
``Locobot: An open source low cost robot,''
  \url{https://locobot-website.netlify.com/}.

\bibitem{sutton2018reinforcement}
R.~S. Sutton and A.~G. Barto, \emph{Reinforcement learning: An
  introduction}.\hskip 1em plus 0.5em minus 0.4em\relax MIT press, 2018.

\bibitem{hector_slam}
S.~Kohlbrecher, J.~Meyer, O.~von Stryk, and U.~Klingauf, ``A flexible and
  scalable slam system with full 3d motion estimation,'' in \emph{SSRR}.\hskip
  1em plus 0.5em minus 0.4em\relax IEEE, November 2011.

\bibitem{Choi_2015_CVPR}
S.~Choi, Q.-Y. Zhou, and V.~Koltun, ``Robust reconstruction of indoor scenes,''
  in \emph{IEEE Conference on Computer Vision and Pattern Recognition (CVPR)},
  2015.

\bibitem{pyrobot2019}
A.~Murali, T.~Chen, K.~V. Alwala, D.~Gandhi, L.~Pinto, \emph{et~al.},
  ``Pyrobot: An open-source robotics framework for research and benchmarking,''
  \emph{arXiv preprint arXiv:1906.08236}, 2019.

\bibitem{habitat_challenge}
``Habitat {C}hallenge 2019 @ {H}abitat {E}mbodied {A}gents {W}orkshop. {CVPR}
  2019,'' \url{https://aihabitat.org/challenge/2019/}.

\bibitem{heusel2017gans}
M.~Heusel, H.~Ramsauer, T.~Unterthiner, B.~Nessler, and S.~Hochreiter, ``Gans
  trained by a two time-scale update rule converge to a local nash
  equilibrium,'' in \emph{Advances in neural information processing systems},
  2017, pp. 6626--6637.

\bibitem{DBLP:SadeghiL17}
F.~Sadeghi and S.~Levine, ``{CAD2RL:} real single-image flight without a single
  real image,'' in \emph{Robotics: Science and Systems XIII, Massachusetts
  Institute of Technology, Cambridge, Massachusetts, USA, July 12-16 2017}.

\bibitem{simopt}
Y.~Chebotar, A.~Handa, V.~Makoviychuk, M.~Macklin, J.~Issac, \emph{et~al.},
  ``Closing the sim-to-real loop: Adapting simulation randomization with real
  world experience,'' 05 2019, pp. 8973--8979.

\bibitem{residualRL}
T.~{Johannink}, S.~{Bahl}, A.~{Nair}, J.~{Luo}, A.~{Kumar}, \emph{et~al.},
  ``Residual reinforcement learning for robot control,'' in \emph{2019
  International Conference on Robotics and Automation (ICRA)}, 2019, pp.
  6023--6029.

\bibitem{Silver2018ResidualPL}
T.~Silver, K.~R. Allen, J.~B. Tenenbaum, and L.~P. Kaelbling, ``Residual policy
  learning,'' \emph{ArXiv}, vol. abs/1812.06298, 2018.

\end{thebibliography}
